\documentclass[conference]{IEEEtran}
\IEEEoverridecommandlockouts
\usepackage{amsmath,amssymb,amsfonts}
\usepackage{algorithmic}
\usepackage{graphicx}
\usepackage{float}
\usepackage{textcomp}
\usepackage{biblatex}
\usepackage{xcolor}
\usepackage{multicol}
\usepackage{subcaption}
\usepackage{longtable}
\usepackage{lipsum}
\usepackage{booktabs, capt-of, tabularx}
\usepackage{cuted, stfloats}
\usepackage{hyperref}
\hypersetup{
    colorlinks=true,
    linkcolor=black,
    filecolor=black,
    citecolor=black,
    urlcolor=black,
}

\begin{document}

\title{Comparative Study of Pre-Trained BERT Models for Code-Mixed Hindi-English Data\\
}

\makeatletter
\newcommand{\newlineauthors}{%
  \end{@IEEEauthorhalign}\hfill\mbox{}\par
  \mbox{}\hfill\begin{@IEEEauthorhalign}
}
\makeatother

\author{\IEEEauthorblockN{Aryan Patil}
\IEEEauthorblockA{
\textit{Pune Institute of Computer Technology}\\
\textit{L3Cube Pune}\\
Thane, India \\
aryanpatil1503@gmail.com}
\and
\IEEEauthorblockN{Varad Patwardhan}
\IEEEauthorblockA{
\textit{Pune Institute of Computer Technology}\\
\textit{L3Cube Pune}\\
Pune, India \\
varadp2000@gmail.com}
\and
\IEEEauthorblockN{Abhishek Phaltankar}
\IEEEauthorblockA{
\textit{Pune Institute of Computer Technology}\\
\textit{L3Cube Pune}\\
Pune, India \\
avp1510@gmail.com}
\and
\newlineauthors
\IEEEauthorblockN{Gauri Takawane}
\IEEEauthorblockA{
\textit{Pune Institute of Computer Technology}\\
\textit{L3Cube Pune}\\
Pune, India \\
gauri.takawane@gmail.com}
\and
\IEEEauthorblockN{Raviraj Joshi}
\IEEEauthorblockA{\textit{Indian Institute of Technology Madaras} \\
\textit{L3Cube Pune}\\
Pune, India \\
ravirajoshi@gmail.com}
}

\maketitle

\begin{abstract}
The term ”Code Mixed” refers to the use of more than one language in the same text. This phenomenon is predominantly observed on social media platforms, with an increasing amount of adaptation as time goes on. It is critical to detect  foreign elements in a language and process them correctly, as a considerable number of individuals are using code-mixed languages that could not be comprehended by understanding one of those languages. In this work, we focus on low-resource Hindi-English code-mixed language and enhancing the performance of different code-mixed natural language processing tasks such as sentiment analysis, emotion recognition, and hate speech identification. We perform a comparative analysis of different Transformer-based language Models pre-trained using unsupervised approaches. We have included the code-mixed models like HingBERT, HingRoBERTa, HingRoBERTa-Mixed, mBERT, and non-code-mixed models like AlBERT, BERT, and RoBERTa for comparative analysis of code-mixed Hindi-English downstream tasks. We report state-of-the-art results on respective datasets using HingBERT-based models which are specifically pre-trained on real code-mixed text. Our HingBERT-based models provide significant improvements thus highlighting the poor performance of vanilla BERT models on code-mixed text.

\end{abstract}


\section{Introduction}
\begin{table*}[!h]
\section{Literature Review}\label{sec:Literature Review}
\begin{center}
\begin{tabular}{|l|p{0.27\linewidth}|l|p{0.27\linewidth}|p{0.28\linewidth}|}
\hline
\textbf{}&\multicolumn{4}{|c|}{\textbf{Summary of benchmark research}}\\
\cline{2-5}
\textbf{Sr.No}&\textbf{Paper Title/ Author}&\textbf{Year}&\textbf{Summary}&\textbf{Limitations} \\
\hline
1 & Corpus Creation for Sentiment Analysis in Code-Mixed Tamil-English Text Bharathi Raja Chakravarthi,Vigneshwaran Muralidaran, Ruba Priyadharshini, John P. McCrae \textsuperscript{\cite{DBLP:journals/corr/abs-2006-00206}} & 2022 & They created corpora consisted of Tamil-English code-switched YouTube comment posts which were sentiment-annotated. & Using linguistic data or hierarchical meta-embedding can further improve code-mixed models by reducing the level of supervision and expanding the context across the entire dataset.\\
\hline
2 & A Sentiment Analysis Dataset for Code-Mixed Malayalam-English Bharathi Raja Chakravarthi, Navya Jose, Shardul Suryawanshi, Elizabeth Sherly, John P. McCrae \textsuperscript{\cite{DBLP:journals/corr/abs-2006-00210}} & 2020 & It provides a new standard corpus for sentiment analysis of code-mixed text in Malayalam-English and is dedicated to establishing standards for sentiment analysis. & The corpus is solely made up of YouTube comments, and the BERT model only works with English text, which might impede Malayalam-English data .\\
\hline
3 & Detection of Hate Speech Text in Hindi-English Code-mixed Data Sreelakshmi K, Premjith B, Soman K.P. \textsuperscript{\cite{SREELAKSHMI2020737}} & 2020 & It developed a ML model to identify hate speech in the code-mixed text due to colloquial spelling \& grammar differences \& blending of different languages. & Hate speech can be further classified as Offensive, Abusive, and Profane. \\
\hline
4 & Towards Sub-Word Level Compositions for Sentiment Analysis of Hindi-English Code Mixed Text Ameya Prabhu, Aditya Joshi, Manish Shrivastava and Vasudeva Varma \textsuperscript{\cite{DBLP:journals/corr/PrabhuJSV16}} & 2016 &This research presents a new approach for learning sub-word level representations in LSTM architecture, which improves performance in text with excessive noise. A dataset for sentiment analysis is also provided as a contribution. & The proposed technique has the potential to be applied to larger architectures and the dataset can also be expanded. Furthermore, the research could also be done on deep neural network architectures. \\
\hline
5 & Humor Detection in English-Hindi Code-Mixed Social Media Content : Corpus and Baseline System Ankush Khandelwal, Sahil Swami, Syed S. Akhtar, Manish Shrivastava \textsuperscript{\cite{DBLP:journals/corr/abs-1806-05513}} & 2018 & It presents a dataset of social media texts in code-mixed Hindi-English that has been annotated for humor detection and word-level language tagging. & The corpus can be annotated by parts of speech tags for better results and the dataset can be extended by adding tweets from other domains. \\
\hline
6 & Sentiment Analysis of Code-Mixed Indian Languages: An Overview of SAIL Code- Mixed Shared Task @ICON-2017 Braja Gopal Patra, Dipankar Das, and Amitava Das \textsuperscript{\cite{Patra2018SentimentAO}} & 2018 & The shared task that was undertaken was the sentiment analysis of code-mixed data pairs, which were tweets in Hindi-English and Bengali-English collected from Twitter users in the Indian subcontinent. & Due to a shortage of sufficient datasets, conventional machine learning models have outperformed deep learning models. Also, more complex tasks like aspect-based sentiment analysis and stance detection can be executed. \\
\hline 
7 & Towards Emotion Recognition in Hindi-English Code-Mixed Data: A Transformer Based Approach Anshul Wadhawan and Akshita Aggarwal \textsuperscript{\cite{https://doi.org/10.48550/arxiv.2102.09943}} & 2021 & This paper contribute by offering a labeled dataset for experiments on emotion detection in Hindi-English tweets using both deep learning and transformer-based approaches. & By comparing MUSE-aligned vectors, pre-aligned FastText word embedding, and language-specific transformer-based word embeddings, better results can be achieved. Improve handling of linguistic complexities and preprocessing for cleaner data. \\
\hline 
8 & A Corpus of English-Hindi Code-Mixed Tweets for Sarcasm Detection Sahil Swami , Ankush Khandelwal , Vinay Singh , Syed Sarfaraz Akhtar and Manish Shrivastava \textsuperscript{\cite{DBLP:journals/corr/abs-1805-11869}} & 2018 & It provides a dataset annotated with both sarcasm and language tagging and provides a baseline supervised classification performed using the same dataset with10-fold cross validation using machine learning approaches. & The dataset is skewed. The use of word embeddings, POS tags, and language-based features can further enhance this system.\\
\hline
9 & Hate Speech Detection from Code-mixed Hindi-English Tweets Using Deep Learning Models Satyajit Kamble, Aditya Joshi \textsuperscript{\cite{DBLP:journals/corr/abs-1811-05145}} & 2018 & It has examined the use of domain-specific word embeddings for hate speech detection using deep learning models. & It could include work on variations of language with sarcasm or misinformation. Deep learning approaches that can correctly assimilate textual signals can be applied .\\
\hline
10 & Part-of-Speech Tagging for Code-Mixed English-Hindi Twitter and Facebook Chat Messages Anupam Jamatia, Björn Gambäck and Amitava Das \textsuperscript{\cite{Jamatia2015PartofSpeechTF}} & 2015 & 
This study tries to capture and analyze social media information on Hindi-English code mixed text along with an attempt to automate the parts-of-speech tagging process. & Language modeling on code-mixed text  can be researched to solve the issues brought on by unidentified words.\\
\hline
11 & De-Mixing Sentiment from Code-Mixed Text Yash Kumar Lal, Vaibhav Kumar, Mrinal Dhar, Manish Shrivastava and Philipp Koehn \textsuperscript{\cite{lal-etal-2019-de}} & 2019 & In this paper, a hybrid architecture for sentiment analysis of data with English-Hindi code mixture is presented using recurrent neural networks. & By altering the attention processes and training the model on a larger dataset, the model can be assessed.\\
\hline
12 & A Dataset of Hindi-English Code-Mixed Social Media Text for Hate Speech Detection Aditya Bohra, Deepanshu Vijay, Vinay Singh, Syed S. Akhtar, and Manish Shrivastava \textsuperscript{\cite{bohra-etal-2018-dataset}} & 2018 & This study explores the issue of identifying hate speech in code-mixed texts, provides a dataset of tweets that are code-mixed in Hindi and English and introduces a supervised method for hate speech detection. &The corpus lacks word-level part-of-speech annotations. The aforementioned experiments can be expanded to include several languages.\\
\hline

\end{tabular}
\caption{Literature Survey Table}
\label{tab:Survey}
\end{center}
\end{table*}

Text that alternates between two or more languages is referred to as code mixed data. The code-mixed text does not follow any grammatical structure and might have abrupt language transitions \cite{sitaram2019survey,joshi2022evaluating}. Natural Language Processing (NLP) research has been mainly concentrated on monolingual data as compared to code-mixed data. Moreover, code-mixed data is mostly restricted to social media platforms thus limiting the data availability \cite{das2015code}.

It has been noted that over the last few years, there has been a dramatic increase in the use of English, in combination with other regional languages. Like many other non-English speakers worldwide, people from India refrain from using a single code in their online communications. As 'Hindi' is the national language of India and is the most commonly spoken language, code-mixing of Hindi-English text on Social Media is quite frequently found.

This code-mixed form of data is being utilized more frequently and is becoming more ubiquitous in social media chats and in other private chat platforms. There is an apparent mismatch between the scale at which this code-mixed language is used and the amount of data that is currently available for further research \cite{jose2020survey,nayak-joshi-2022-l3cube}.

Building and training a new model from scratch can be tedious as it requires a large amount of data and is both time and energy-consuming. As pre-trained models have already been trained and optimized, a small amount of finetuning data is necessary to build competitive models. 

The BERT-based architectures are the standard pre-trained language models currently used in literature \cite{DBLP:journals/corr/abs-1810-04805}. A lot of studies have been proposed around pre-training and fine-tuning them on various tasks. There have been variations around the BERT architecture like RoBERTa and ALBERT, which have helped in various use cases like accuracy and latency-related improvements. Models like multilingual BERT have focused mainly on multilingual and cross-lingual data representations.

The vanilla BERT models have only been trained on monolingual data with no consideration for code-mixed text. Recently introduced HingBERT-based models are trained on Hindi-English code mixed data and are shown to provide state-of-the-art results on code-mixed benchmark datasets. These models have an advantage over regular models because they are capable of understanding a real code-mixed text. Some of the models are trained on both Roman and Devanagari scripts. 

In this work, we present a comparative study of different BERT models on downstream code-mixed text classification datasets. The BERT models considered in this work include English-based BERT, RoBERTa, and AlBERT, multi-lingual based mBERT, and code-mixed based HingBERT, HingRoBERTa, and HingRoBERTa-Mixed \cite{nayak-joshi-2022-l3cube}. These models are evaluated on five different datasets including two sentiment analysis datasets \cite{Patra2018SentimentAO,DBLP:journals/corr/PrabhuJSV16}, one emotion recognition \cite{https://doi.org/10.48550/arxiv.2102.09943}, and two hate speech datasets \cite{bohra-etal-2018-dataset,lal-etal-2019-de}. We report new state-of-the-art performance on these datasets using HingBERT based models.

The paper is structured as follows: Section \ref{sec:Literature Review} includes information about related work in this domain. Section \ref{sec:Methodologies} describes the datasets and models that were used, as well as the methodology used for data pre-processing and model training. Section \ref{sec:Result} provides the details about the results, and Section \ref{sec:Conclusion} consists of conclusions based on the results.

\section{Methodologies}\label{sec:Methodologies}
The study aims to provide a thorough comparison of Code-Mixed Hindi-English language models with their original counterparts. This section outlines the series of processes that make up the suggested technique including dataset descriptions, data preprocessing methods, and model training.

\subsection{Datasets}

\begin{table}[!h]
\begin{center}
\begin{tabular}{|p{0.03\linewidth}|p{0.15\linewidth}|p{0.08\linewidth}|p{0.08\linewidth}|p{0.07\linewidth}|p{0.07\linewidth}|p{0.08\linewidth}|}
    \hline
    \textbf{Sr. No} & \textbf{Dataset Name} & \textbf{Total size} & \textbf{Train Size} & \textbf{Test Size} & \textbf{Eval Size} & \textbf{Labels}\\
    \hline
    1 & Icon Dataset & 18461 & 12936 & 5525 & 5525 & 3\\
    \hline
    2 & Emotions Dataset & 151311 & 105917 & 22697 & 22697 & 6\\
    \hline
    3 & Sentiment Dataset & 3879 & 2482 & 776 & 621 & 3\\
    \hline
    4 & Hatespeech Dataset &4578 & 3204 & 687 & 687 & 2 \\
    \hline
    5 & HASOC Dataset & 4864 & 3404 & 730 & 730 & 2 \\
    \hline
\end{tabular}
\caption{Dataset Train-Test-Validation Split}
\label{tab:Datasets}
\end{center}
\end{table}

Datasets are gathered from five distinct sources. The dataset information is provided as follows -\\
\subsubsection{Icon Dataset \textsuperscript{\cite{Patra2018SentimentAO}}}
The Icon shared task served as the source of this dataset. It is made up of Hindi-English code-mix tweets. It has 12936 training samples and 5525 test samples. There are three emotions present namely positive, negative, and neutral. The Icon shared task's overall baseline f1 score was 0.331, and the best model's f1 score came from IIIT-NBP, where it was 0.569. 

\subsubsection{Emotions Dataset \textsuperscript{\cite{https://doi.org/10.48550/arxiv.2102.09943}}}
This dataset was taken from a study that focused on emotion recognition. Six categories have been used to group the tweets. There are altogether 151311 tweets. A total of 26364 tweets express happiness, 21024 tweets express sadness, 29306 tweets express anger, 19138 tweets express fear, 35797 tweets express disgust, and 19682 tweets express surprise. The BERT model produced the highest accuracy score, which was around 71.43\%. In a ratio of 70:15:15, the dataset is divided into three sections. For training, we are using 70\% of the dataset. 15\% for validating and for testing the remaining 15\%.\\

\subsubsection{Sentiment Dataset \textsuperscript{\cite{DBLP:journals/corr/PrabhuJSV16}}}
This dataset is proposed in paper \cite{DBLP:journals/corr/PrabhuJSV16}. The sentiment of tweets is divided into three categories namely negative, neutral, and positive. It consists of 3879 tweets in total. The best accuracy score and f1 score using the Subword-LSTM method are 69.7\% and 0.658, respectively. The dataset is split in an 80:20 ratio. Testing is done on 20\% of the dataset. Once more, an 80:20 ratio is used to distribute the remaining 80\%. The models will be trained using 80\% of this data, and the remaining 20\% will be utilized for validation.\\

\subsubsection{Hatespeech Dataset \textsuperscript{\cite{bohra-etal-2018-dataset}}}
This dataset was curated from a study focused on hate speech detection in Hindi-English code-mixed data. The corpus is classified into two classes - hate speech and normal speech. Out of the 4575 tweets in total, 1661 tweets depict hate speech and 2914 tweets depict normal speech. Support Vector Machines have produced the highest accuracy score 71.7\% when all features are included. In a ratio of 70:15:15, the dataset is divided into three sections. For training, we are using 70\% of the dataset. 15\% for validating and for testing the remaining 15\%.\\

\subsubsection{HASOC Dataset}
This dataset is curated from HASOC Shared Task 2022 which consists of two classes - hate speech and non-hate speech. Out of the 4864 tweets in total, 2514 tweets represent hate speech and 2390 tweets depict non-hate speech. The best F1 score is 0.708 from NLPLAB-ISI. In a ratio of 70:15:15, the dataset is divided into three sections. For training, we are using 70\% of the dataset. 15\% for validating and for testing the remaining 15\%.\\

\begin{figure}[!h]
    \centering
    \includegraphics[scale=0.45]{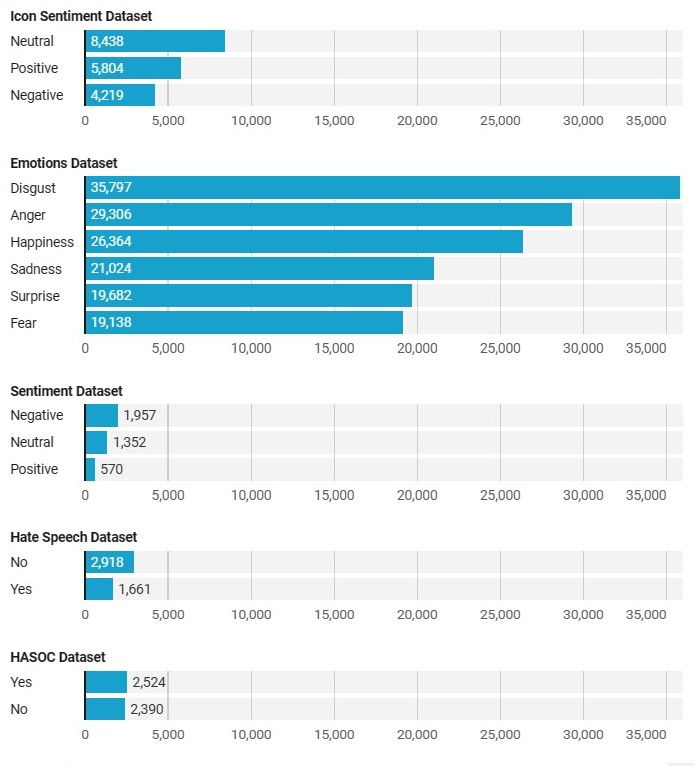}
    \caption{Datasets Distribution}
    \label{fig:Datasets}
\end{figure}

\begin{figure*}[b]
    \centering
    \includegraphics[scale=0.32]{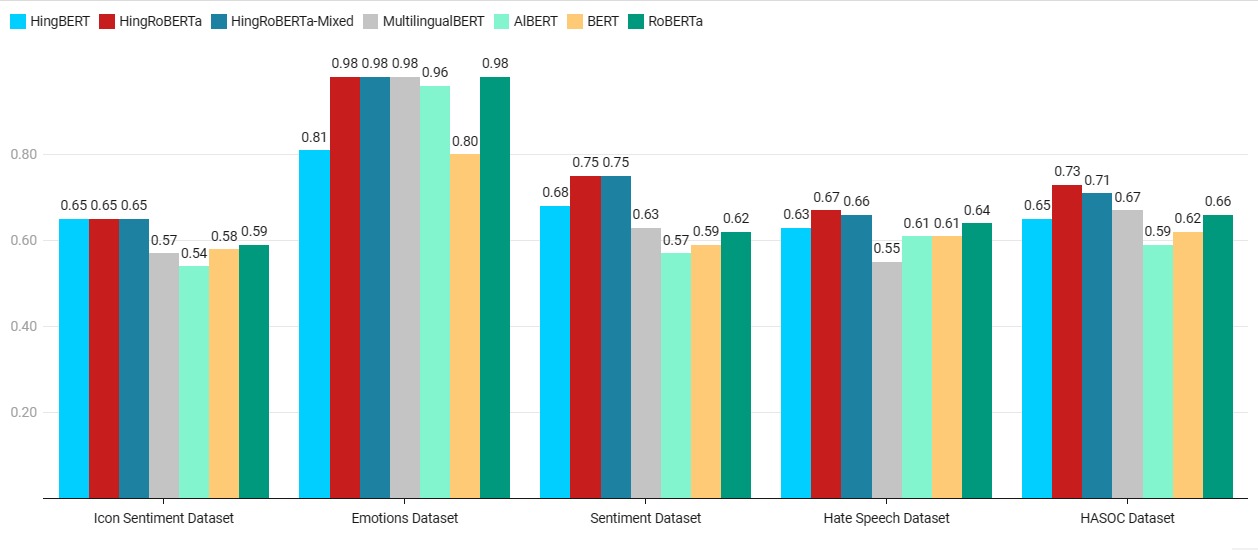}
    \caption{Results of Deep Learning Models (F1 Score - Macro)}
    \label{fig:Results}
\end{figure*}

\subsection{Preprocessing Techniques}
We have used a variety of pre-processing methods to sanitize the text. We removed hyperlinks, username mentions, hashtags, punctuations, newline characters, and spaces using the regex library. We utilized the emoji library to remove the emojis. These preprocessing techniques will help to provide cleaner data for the models.

\subsection{Models} 
We used the following models in our research:-
\subsubsection{HingBERT\textsuperscript{\cite{nayak-joshi-2022-l3cube}}}
A BERT model called HingBERT\footnote{\url{https://huggingface.co/l3cube-pune/hing-bert}} was developed using roman text with both Hindi and English codes mixed in. It is an improved version of the mBERT model on the L3Cube-HingCorpus. 52.93 million phrases from Twitter were used to create the 1.04 billion tokens that made up the HingCorpus.\\

\subsubsection{HingRoBERTa\textsuperscript{\cite{nayak-joshi-2022-l3cube}}}
The HingRoBERTa\footnote{\url{https://huggingface.co/l3cube-pune/hing-roberta}} is an English RoBERTa model fine-tuned on roman Hindi-English code mixed text. Again, roman version of L3Cube-HingCorpus was used for finetuning.\\

\subsubsection{HingRoBERTa-Mixed\textsuperscript{\cite{nayak-joshi-2022-l3cube}}}
The Hindi-English BERT model HingRoBERTa-Mixed\footnote{\url{https://huggingface.co/l3cube-pune/hing-roberta-mixed}} was developed using roman and Devanagari code-mixed text. It's a xlm-RoBERTa model fine-tuned on mixed script (Roman + Devanagari) L3Cube-HingCorpus. \\

\subsubsection{BERT\textsuperscript{\cite{DBLP:journals/corr/abs-1810-04805}}}
The Bidirectional Encoder Representations from Transformers (BERT) model consists of self-supervised transformers which were pre-trained on a sizable corpus of multilingual data comprising of Toronto Book Corpus and Wikipedia. It was trained on raw text, without human labeling using masked language modeling and next sentence prediction task.\\

\subsubsection{Multilingual BERT\textsuperscript{\cite{https://doi.org/10.48550/arxiv.1810.04805}}}
A BERT version called Multilingual Bidirectional Encoder Representations from Transformers (mBERT) was trained on the large multi-lingual dataset employing 104 different languages.  \\

\subsubsection{ALBERT\textsuperscript{\cite{DBLP:journals/corr/abs-1909-11942}}}
For self-supervised learning of language representations, a Lite BERT (ALBERT) model was presented. This model's architectural foundation is similar to that of BERT. The main goal was to increase the performance of the BERT model by applying various design decisions to decrease the number of training parameters making it memory and time efficient. \\

\subsubsection{RoBERTa\textsuperscript{\cite{DBLP:journals/corr/abs-1907-11692}}}
Robustly Optimized BERT Pretraining Approach is known as RoBERTa. It is again based on Google's BERT model published in 2018. Due to the removal of the next-sentence pretraining and the use of considerably bigger batches, the training is quicker and more effective than BERT.\\

\subsection{Methodology}
For training these models we have utilized multiple datasets which are mentioned in the paper. We have implemented the above models to find the accuracy, precision, recall, and F1 score of the datasets. We are using simple transformers as a wrapper to tokenize and encode the text. After that we are using WandB\textsuperscript{\cite{wandb}} for hyperparameter tuning. We are adjusting the range of the learning rate between 1e-6 and 1e-4. We are adjusting the range of epochs between 1 to 5. We are adjusting the batch size between 32 and 64. And we are training the model on the best parameters to find and compare the results.\\

\begin{table*}[!h]
\begin{center}
\begin{tabular}{|c|c|c|c|c|c|c|c|c|}
\hline
\textbf{Dataset}&\textbf{Model}&\multicolumn{3}{|c|}{\textbf{Parameters}}&\multicolumn{4}{|c|}{\textbf{Scores}} \\
\cline{3-9} 
\textbf{Used} &\textbf{Used} & \textbf{\textit{Learning Rate}}& \textbf{\textit{Epochs}}& \textbf{\textit{Batch Size}} & \textbf{\textit{F1 score}}& \textbf{\textit{Precision}}& \textbf{\textit{Recall}}&\textbf{\textit{Accuracy}} \\
\hline \hline
\textbf{} & HingBERT & 9.1974E-05 & 1 & 64 & 0.62825 & 0.64530 & 0.62069 & 0.64633\\
\cline{2-9} 
\textbf{} & HingRoBERTa & 1.1047E-05 & 4 & 32 & 0.65222 & 0.65765 & 0.64793 & 0.66407\\
\cline{2-9}
\textbf{} & HingRoBERTa-Mixed & 4.8418E-05 & 3 & 64 & \textbf{0.65288} & 0.65711 & 0.64940 & 0.66425\\
\cline{2-9}
Icon Hi-En & Multilingual BERT &  7.2483E-05 & 3 & 32 & 0.56593 & 0.57110 & 0.56441 & 0.58643\\
\cline{2-9}
\textbf{} & Albert & 6.1757E-05 & 4 & 64 & 0.54280 & 0.56233 & 0.53655 & 0.57756\\
\cline{2-9}
\textbf{} & BERT & 5.8090E-05 & 2 & 32 & 0.58004 & 0.60080 & 0.57166 & 0.61104\\
\cline{2-9}
\textbf{} & RoBERTa & 4.0715E-05 & 4 & 32 & 0.58730 & 0.59607 & 0.58205 & 0.61050\\
\hline \hline
\textbf{} & HingBERT & 9.1974E-05 & 1 & 64 & 0.81437 & 0.81890 & 0.81562 & 0.82381\\
\cline{2-9} 
\textbf{} & HingRoBERTa &  1.1047E-05 & 4 & 32 & 0.98195 & 0.98177 & 0.98215 & 0.98401\\
\cline{2-9} 
\textbf{} & HingRoBERTa-Mixed & 4.8418E-05 & 3 & 64 & \textbf{0.98285} & 0.98267 & 0.98306 & 0.98480\\
\cline{2-9} 
Emotions & Multilingual BERT & 7.2483E-05 & 3 & 32 & 0.97705 & 0.97587 & 0.97841 & 0.97960\\
\cline{2-9} 
\textbf{} & Albert & 6.1757E-05 & 4 & 64 & 0.95540 & 0.95502 & 0.95619 & 0.95867\\
\cline{2-9} 
\textbf{} & BERT & 5.8090E-05 & 2 & 32 & 0.80423 & 0.80510 & 0.80702 & 0.81271\\
\cline{2-9} 
\textbf{} & RoBERTa & 4.0715E-05 & 4 & 32 & 0.98001 & 0.97918 & 0.98091 & 0.98264\\
\hline \hline
\textbf{} & HingBERT & 2.2572E-05 & 4 & 32 & 0.67471 & 0.68192 & 0.66963 & 0.69845\\
\cline{2-9} 
\textbf{} & HingRoBERTa & 1.1047E-05 & 4 & 32 & 0.74705 & 0.75763 & 0.73833 & 0.75515\\
\cline{2-9} 
\textbf{} & HingRoBERTa-Mixed & 8.2990E-05 & 2 & 64 & \textbf{0.74898} & 0.74997 & 0.74937 & 0.76160\\
\cline{2-9} 
Sentiment & Multilingual BERT & 7.9518E-05 & 5 & 32 & 0.62456 & 0.62454 & 0.62471 & 0.65464\\
\cline{2-9} 
\textbf{} & Albert & 5.3669E-05 & 5 & 64 & 0.57018 & 0.57840 & 0.56850 & 0.61211\\
\cline{2-9} 
\textbf{} & BERT & 6.4191E-05 & 4 & 32 & 0.59549 & 0.60492 & 0.59056 & 0.63273\\
\cline{2-9} 
\textbf{} & RoBERTa & 7.6186E-05 & 5 & 32 & 0.62381 & 0.63988 & 0.61984 & 0.66237\\
\hline \hline
\textbf{} & HingBERT & 2.6585E-05 &2 &32 &0.62860 &0.65713 &0.62398 &0.69432\\
\cline{2-9} 
\textbf{} & HingRoBERTa & 4.1386E-05 &5 &64 &\textbf{0.66994} &0.68239 &0.66425 &0.71470\\
\cline{2-9} 
\textbf{} & HingRoBERTa-Mixed & 8.3211E-05 & 5 & 32 & 0.65990 & 0.68836 & 0.65266 & 0.71761\\
\cline{2-9} 
 Hatespeech& Multilingual BERT & 3.8620E-05 & 5 & 64 & 0.54588 & 0.68805 & 057017. & 0.68850\\
\cline{2-9} 
\textbf{} & Albert & 9.3810E-05 & 5 & 64 & 0.61399 & 0.69097 & 0.61426 & 0.70742\\
\cline{2-9} 
\textbf{} & BERT & 3.4398E-05 & 3  & 32 & 0.60958 & 0.71431 & 0.61279 & 0.71325\\
\cline{2-9} 
\textbf{} & RoBERTa & 1.9650E-05 & 4 & 64 & 0.63983 & 0.71549 & 0.63539 & 0.72344\\
\hline \hline
\textbf{} & HingBERT & 4.7456E-05& 3 & 64 & 0.65204 & 0.65205 & 0.65204 & 0.65205\\
\cline{2-9} 
\textbf{} & HingRoBERTa & 4.8000E-05 & 4 & 64 & \textbf{0.73270} & 0.73337 & 0.73281 & 0.73287\\
\cline{2-9} 
\textbf{} & HingRoBERTa-Mixed & 6.8471E-05 & 2 & 64 & 0.71087 & 0.71114 & 0.71091 & 0.71095\\
\cline{2-9} 
HASOC & Multilingual BERT & 0.9998E-05 & 4 & 64 & 0.67193 & 0.67381 & 0.67248 & 0.67260\\
\cline{2-9} 
\textbf{} & Albert & 5.7947E-05 & 4 & 64 & 0.59617 & 0.60141 & 0.59885 & 0.59863\\
\cline{2-9} 
\textbf{} & BERT & 9.3592E-05 & 4 & 64 & 0.61643 & 0.61647 & 0.61646 & 0.61644\\
\cline{2-9} 
\textbf{} & RoBERTa & 1.0247E-05 & 5 & 32 & 0.66259 & 0.66369 & 0.66292 & 0.66301\\
\hline
\end{tabular}
\caption{Results of Deep Learning Models}
\label{tab:Results}
\end{center}
\end{table*}

\section{Result}\label{sec:Result}
The accuracy, recall, precision, and f1score of the five datasets listed above using the following deep learning models - HingBERT, HingRoBERTa, HingRoBERTa-Mixed, Multilingual BERT, ALBERT, BERT, and RoBERTa are shown in Table \ref{tab:Results}. We can rule out the possibility of overfitting a model since the accuracy and recall scores are good.


Table \ref{tab:Results} shows that HingRoBERTa and HingRoBERTa-Mixed consistently beat all other models in terms of f1score, with HingBERT being the second-best model. ALBERT performs the worst out of all of them, as predicted, while BERT, RoBERTa, and multilingual BERT all do very well. As they were all trained on romanized Hindi English code mix data, HingBERT, HingRoBERTa, and HingRoBERTa-Mixed outperform base BERT and RoBERTa, which were only trained in English. Specifically, HingRoBERTa performs the best on Icon, Hate speech, and HASOC datasets while HingRoBERTa-Mixed provides state-of-the-art results for the Emotions and Sentiment dataset.

Multilingual BERT is not trained on code mix data but was trained on Devnagari script languages like Hindi and Marathi. Consequently, it works rather well even if it isn't as good as the Hing models. Since it was trained using fewer data than BERT or RoBERTa and just on the English dataset, ALBERT, a lighter variant of BERT, performs the poorest. The f1-score and accuracy from basic models were improved by the hyper tuning carried out with WandB\textsuperscript{\cite{wandb}}.

\section{Conclusion}\label{sec:Conclusion}
Sentiment analysis and opinion mining have emerged as major fields of research due to the increase in recent years in the use of social media for the free expression of ideas.

In this study,  we employ BERT-based deep learning models to enhance the accuracy and f1scores of already existing code-mixed Hindi and English Twitter datasets with labels composed of sentiments, emotions, and hate speech. HingBERT-based models perform better than the other deep learning models that have been studied.

This paper summarises our study, which attempts to enhance low-resource Hindi-English models by leveraging corresponding code-mixed BERT models.

In the future, HingBERT models can be further employed for name entity recognition and other language analysis studies. Additionally, code-mixed resources like HingFT and HingGPT also need to be evaluated on classification and generation tasks, respectively. 

\section*{Acknowledgements} 
This work was completed as part of the L3Cube Mentorship Program in Pune. We would like to convey our thankfulness to our L3Cube mentors for their ongoing support and inspiration.

\printbibliography
\end{document}